\documentclass[a4paper]{article}

\usepackage[utf8]{inputenc} 
\usepackage[T1]{fontenc}    
\usepackage{hyperref}       
\usepackage{url}
\usepackage{booktabs}
\usepackage{amsfonts}
\usepackage{xcolor}
\usepackage{siunitx}
\usepackage{eurosym}
\usepackage{amsmath}
\usepackage{amsthm}
\usepackage{algorithm}
\usepackage{algorithmicx}
\usepackage{algpseudocode}
\usepackage{graphicx}
\usepackage{natbib}
\usepackage{geometry}
\usepackage{eurosym}
\usepackage{crimson}

\newcommand{\Do}{\mathrm{do}}
\newcommand{\E}{\mathbb E}
\DeclareMathOperator{\Var}{Var}
\DeclareMathOperator{\Span}{Span}
\DeclareMathOperator{\Dir}{Dir}

\DeclareMathOperator{\Binom}{B}
\DeclareMathOperator{\Cat}{Cat}
\newcommand{\dd}{\,\mathrm d}

\theoremstyle{plain}
\newtheorem{theorem}{Theorem}

\theoremstyle{remark}

\theoremstyle{definition}
\newtheorem{definition}{Definition}

\title{Partial counterfactual identification and uplift modeling: theoretical results and real-world assessment}

\author{
  Théo Verhelst\thanks{
  Machine Learning Group, Department of Computer Science, Université Libre de Bruxelles, Brussels, Belgium}\\
  \texttt{theo.verhelst@ulb.be}
  \and
  Denis Mercier\thanks{
  Data Science Team, Orange Belgium, Brussels, Belgium}\\
  \texttt{denis1.mercier@orange.com}
  \and
  Jeevan Shrestha\footnotemark[2]\\
  \texttt{jeevan.shrestha@orange.com}
  \and
  Gianluca Bontempi\footnotemark[1]\\
  \texttt{gianluca.bontempi@ulb.be}
}
\date{}

\begin{document}

\maketitle

\begin{abstract}
Counterfactuals are central in causal human reasoning and the scientific discovery process.  The uplift, also called conditional average treatment effect,  measures the causal effect of some action, or treatment, on the outcome of an individual. This paper discusses how it is possible to derive bounds on  the probability of counterfactual statements based on uplift terms. First, we derive some original bounds on the probability of counterfactuals and we show that tightness of such bounds depends on the information of the feature set on the uplift term. Then, we propose a point estimator based on the assumption of conditional independence between the counterfactual outcomes. The quality of the bounds and the point estimators are assessed on synthetic data and a large real-world customer data set provided by a telecom company, showing significant improvement over the state of the art.
\end{abstract}


\section{Introduction}
\label{sec:introduction}

Counterfactual statements (or counterfactuals for short) concern the potential of events in situations different from the actual state of the world. An example of counterfactual statement is "I got no effect since I made no action but something would have happened had I acted". Counterfactuals are used in many fields, ranging from algorithmic recourse~\citep{karimi2021algorithmic} to online advertisement and customer relationship management~\citep{li2019unit}. 

Counterfactuals have been formally defined in terms of structural causal models by \citet{pearl2009causality}.
Nevertheless, since a counterfactual statement cannot be directly observed, the research focuses on 
estimating or bounding their probability (e.g. the probability that we have an effect given a treatment and no effect else). 
The probability of some specific counterfactual expressions  have been studied in the literature \citep{tian2000probabilities} because of their relevance in causal decision-making.  The \emph{probability of necessity} (PN) is the probability that an event $y$ would not have occurred in the absence of an action or treatment $t$, given that $y$ and $t$ in fact occurred. Conversely, the \emph{probability of sufficiency} (PS) is the probability that event $y$ would have occurred in the presence of an action $t$, given that both $y$ and $t$ in fact did not occur. Lastly, the \emph{probability of necessity and sufficiency} (PNS) is the probability that the event $y$ occurs if and only if the event $t$ occurs.

In the case of incomplete knowledge about the causal model, identification procedures indicate when and how the probability of counterfactuals can be computed from a combination of observational data, experimental data (i.e. data with randomized treatment), and causal assumptions~\citep{correa2021nested}. In situations where the exact probability of counterfactuals cannot be directly computed, an alternative consists in bounding this quantity. This problem, called \emph{partial counterfactual identification}, has first been addressed by~\cite{tian2000probabilities}, and more recently by~\citet{mueller2021causes} and~\citet{zhang2022partial}.

Counterfactual reasoning has practical applications in business, notably churn modeling: consider a company wishing to use direct marketing actions to prevent customers from churning (i.e. stop using their service). The behavior of the customers in reaction to the two possible actions (contact or not) could be described in terms of counterfactual statements~\citep{devriendt2019why}:
\begin{itemize}
    \item \emph{Sure thing}: customer not churning regardless of the action.
    \item \emph{Persuadable}: customer churning only if not contacted.
    \item \emph{Do-not-disturb}: customer churning only if contacted.
    \item \emph{Lost cause}: customer churning regardless of the action.
\end{itemize}
Note that the probability of a customer being a \emph{do-not-disturb} is  an example of PNS~\citep{tian2000probabilities} while, to the best of our knowledge, the other three probabilities have not been labeled in the causal inference literature. 
Though not observable, those quantities are relevant for adequate decision-making, and partial counterfactual identification can help in reducing the uncertainty
about the possible customer behaviors.

Uplift modeling, where uplift stands for the conditional average treatment effect (CATE), or heterogeneous treatment effect~\citep{zhang2021unified}, is another well-known approach for estimating causal effects. It returns an estimate at the individual level of the impact of some action on the probability of the outcome. In the example of churn prevention, uplift modeling estimates the impact of a promotional offer on the probability of churn for each customer. Most recent and powerful uplift models are based on machine learning~\citep{curth2021nonparametric}. Some uplift models expect experimental data and are based on conventional classification models~\citep{jaskowski2012uplift,athey2016recursive}. Other models accept observational (possibly confounded) data and estimate the uplift through some sort of adjustment, for example with propensity scores in \citep{kunzel2019metalearners} and \citep{curth2021nonparametric}.

Counterfactuals and uplift are closely related, yet formally distinct notions. The counterfactual distribution describes the probability of each possible combination of realized and hypothetical outcomes, while the uplift describes the change in outcome probability due to the treatment. While the counterfactual distribution is more informative, it is also more difficult to estimate than the uplift. In~\citep{li2019unit} it is mentioned that the similarity between these two notions can lead to confusion, especially since they collapse under the assumption of monotonicity (the absence of negative causal effects).

 Existing works on partial counterfactual identification~\citep{mueller2021causes,zhang2022partial} make
structural assumptions on the causal model to derive bounds whose estimate requires a combination of experimental and observational data.
In this paper, we propose some original bounds on the probability of counterfactuals based on the uplift terms. The originality of  our approach consists in defining bounds that depend on terms (like uplift) for which nowadays a lot of 
reliable estimators exist in literature. This is of particular interest in big data applications, where structural assumptions are hard to validate but a large number
of observations about individual descriptors (covariates) and 
past behavior are available.

The main contributions of this paper are as follows:
\begin{itemize}
    \item A set of original bounds on the probability of counterfactuals, expressed in terms of the uplift quantity.
    \item A formal derivation of the relationship between our original bounds and the state-of-the-art Fréchet bounds derived by~\citet{tian2000probabilities}.
    \item A point estimator of the counterfactual probabilities based on the conditional independence assumption.
    \item A hierarchical Bayesian model for simulating counterfactual settings and assessing the accuracy of the sample version of the derived bounds.
    \item A real-world assessment of the proposed bounds with a large data set of customer churn campaigns and a discussion of the potential benefits.
\end{itemize}

The rest of this paper is organized as follows. In Section~\ref{sec:related_work}, we present related work in the literature on partial counterfactual identification. In Section~\ref{sec:formalism}, we present the formalism used throughout this paper. In Sections~\ref{sec:bounds_cf} and \ref{sec:point_cf}, we derive bounds and point estimates on the probability of counterfactuals. We analyze the behavior of these estimators under various conditions with simulated examples in Section~\ref{sec:simulated}. We apply our estimator to a real-world data set from our industrial partner and estimate the suggested potential benefits in Section~\ref{sec:empirical}. Conclusions and limitations are given in Section~\ref{sec:conclusion}.

\section{Related work}
\label{sec:related_work}

The \emph{probability of necessity and sufficiency} (PNS) as presented by~\citet[p. 286]{pearl2009causality} is one of the four counterfactual probabilities that we consider in this paper. Seminal works on partial counterfactual identification include \citep{balke1994counterfactual} and \citep{tian2000probabilities}. They show that in the exogenous case (e.g. when the treatment is randomly assigned), the bounds on the PNS reduce to the Fréchet bounds~\citep{frechet1935generalisation}. We will use these bounds as a baseline in the remaining of this paper.

The PNS conditioned on a set of covariates $x$ is called $x$-specific PNS in \citep{li2019unit}. The main focus of~\citet{li2019unit} is the estimation of the benefit generated by a customer retention campaign when the different types of customers have different values. For example, keeping a persuadable customer (a customer who does not churn only when targeted) might be more beneficial than keeping a customer who would never leave, besides the cost of the targeted action.
In \citep{li2022unit}, the authors further refine the bounds on the campaign benefit based on causal assumptions derived from causal diagrams.

\citet{mueller2021causes} derived tighter bounds on the PNS for a variety of causal diagrams, such as with sufficient covariates or with a mediator variable. In particular, the bounds in Theorem~5 in \citep{mueller2021causes} are formally very close to the bounds we develop in this paper, although they consider a set of discrete covariates, whereas we use uplift modeling which allows for arbitrary high-dimensional covariate sets. \citet{zhang2022partial} express the problem of bounding the probability of counterfactuals into polynomial programming, providing tight bounds for any causal graph and combination of experimental and observational data.

Our approach in this paper differs from \citet{mueller2021causes} and \citet{zhang2022partial} in that we make very few causal assumptions (only that the treatment is randomized), but we suggest uplift modeling as a powerful way to estimate conditional probabilities, and we analyze the impact of mutual information between the conditioning set and the potential outcomes.

\section{Notation}
\label{sec:formalism}
In this section, we present the mathematical notation used throughout this paper. A summary is given in Table~\ref{tab:notation}.

\begin{table}
    \begin{center}
    \begin{minipage}{0.95\textwidth}
    \caption{Mathematical notation.}\label{tab:notation}%
    \begin{tabular}{ll}
        \toprule
        $Y\in\{0,1\}$ & Outcome \\
        $T\in\{0,1\}$ & Treatment indicator \\
        $X$ & Set of features \\ 
        $\Do(T=t)$ & Intervention $T=t$ \\
        $P(Y_t=y)$ & Probability of the outcome $Y=y$ under $\Do(T=t)$ \\
        $S_0,S_1$ & $P(Y_0=1), P(Y_1=1)$ \\
        $\alpha,\beta,\gamma,\delta$ & $P(Y_0=0,Y_1=0),\dots,P(Y_0=1,Y_1=1)$ (see Equations~\eqref{eq:alpha} to~\eqref{eq:delta})\\
        $Q(x)$ & Quantity $Q$ conditioned on $X=x$, e.g. $S_0(x)=P(Y_0=1\mid X=x)$ \\
        \bottomrule
    \end{tabular}
    \end{minipage}
    \end{center}
\end{table}

We use Pearl's causal framework, which is based on the notion of \emph{structural causal models} (SCM). A formal definition of SCMs is given by~\citet[Def. 7.1.1]{pearl2009causality}. In this framework, $T$ denotes the action or \emph{treatment}, $Y$ the causal effect (or \emph{outcome}), $X$ a set of  features (or \emph{covariates}) describing the unit/individual under treatment and the $\Do(T=t)$ operator denotes a causal intervention in the system. In this paper, we will limit ourselves to consider binary treatments and outcomes. For example, let $T$ be the binary variable representing a medical treatment: the notation $\Do(T=1)$ indicates that the treatment is forced on an individual regardless of whether they would have received it without explicit intervention. The conditional probability of $Y=y$ given $X=x$ under the intervention $\Do (T=t)$ is written $P(Y = y\mid \Do(T=t), X=x)$. An alternative notation consists in indicating the intervened variable as a subscript to the other variables, such as\footnote{The features $X$ should also receive the subscript $t$ under the intervention $\Do(T=t)$. In our case, the treatment is supposed to occur after the measurement of $X$, and has thus no effect on $X$. This implies that $X_t = X$.} $P(Y_t=y\mid X=x)$. In our application about customer churn prevention, $Y = 1$ indicates that the customer churned, $X$ a set of descriptive features of the customer and the treatment $T$ denotes the exposure of the customer to a targeted marketing action in the form of an e-mail or a phone call ($T=1$ when targeted, $T=0$ otherwise).

We note the probability of the outcome $Y$ under intervention $\Do(T=0)$ given some feature $X=x$ as
\begin{equation}
    \label{eq:S_0}
    S_0(x)=P(Y_0=1\mid X=x).
\end{equation}
Similarly, under the intervention $\Do(T=1)$ we have
\begin{equation}
    \label{eq:S_1}
    S_1(x)=P(Y_1=1\mid X=x).
\end{equation}
The \emph{uplift} is defined to be the difference between these probabilities:
\begin{equation}
    U(x) = S_0(x) - S_1(x).
\end{equation}
Note that the uplift is also sometimes defined as $U(x)=S_1(x)-S_0(x)$, depending on the context and the meaning of the outcome $Y$. Throughout this paper, the argument $x$ in quantities such as $S_0(x)$ indicates the conditioning on $X = x$. If omitted, the quantity is supposed to be no longer conditioned on $x$ (e.g. $S_0=P(Y_0=1)$). Equivalently, we can consider $S_0(x)$ as a function from the domain of $X$ to $[0,1]$, therefore we can define $S_0$ as $S_0=\mathbb E_X[S_0(X)]$, and similarly for $S_1$ and $U$.

The probabilities $S_0(x)$ and $S_1(x)$ cannot be estimated without further assumptions. In this paper, we make the assumption of \emph{unconfoundedness}\footnote{Also called \emph{ignorability} by \citet{rosenbaum1983central}, \emph{exogeneity} by \citet{pearl2009causality}, or \emph{conditional independence assumption} by \citet{gutierrez2016causal}.}~\citep[Def. 9.2.9]{pearl2009causality}:

\begin{definition}[Unconfoundedness]
\label{def:unconfoundedness}
A variable $Y$ is \emph{unconfounded} with respect to $T$ given $X$ if, for any values $y$, $t$ and $x$,
\[P(Y=y\mid\Do(T=t),X=x)=P(Y=y\mid T=t,X=x)\]
Or, alternatively, if for any value $t$,
\[Y_t\perp T\mid X.\]
\end{definition}
Note that in~\citep{pearl2009causality}, the unconfoundedness is called \emph{exogeneity} and is defined without conditioning on $X$. The distinction is made between \emph{weak} exogeneity and \emph{strong} exogeneity: Definition~\ref{def:unconfoundedness} corresponds to weak exogeneity, while strong exogeneity assumes $\{Y_0,Y_1\}\perp T\mid X$. This distinction has no impact on the results presented in this paper.

Unconfoundedness allows the estimation of the scores $S_0(x)$ and $S_1(x)$ from data, since \[S_0(x)=P(Y=1\mid\Do(T=0),X=x)=P(Y=1\mid T=0,X=x)\] and similarly for $S_1(x)$. Unconfoundedness is guaranteed when the treatment $T$ is randomized. In absence of randomization by using a suitable adjustment set (i.e. satisfying the \emph{back-door criterion}~\citep{pearl2009causality}) an estimation method could still permit the unbiased estimation of $S_0(x)$ and $S_1(x)$. Such assumption is typically made in uplift approaches integrating propensity scores, notably the X-learner~\citep{kunzel2019metalearners} and, more recently, double machine learning estimators~\citep{jung2021estimating}.

Let us suppose that $Y_0=1$, i.e. we observe $Y=1$ after
having assigned the treatment $T=0$ to a given individual. Though we cannot observe the counterfactual outcome $Y_1$, we can reason about the value it would have. If $Y_1=0$, the treatment would have a causal impact on the outcome, since the outcome $Y$ changes by intervening on $T$. Otherwise, if $Y_1=1$, the treatment would have no causal influence on the outcome of this individual. More generally, the joint values of $Y_0$ and $Y_1$ define four different counterfactual expressions. In this paper their probability is noted
\begin{align}
        \alpha &= P(Y_0=0, Y_1=0) \label{eq:alpha} \\       
        \beta  &= P(Y_0=1, Y_1=0) \label{eq:beta}\\
        \gamma &= P(Y_0=0, Y_1=1) \label{eq:gamma}\\
        \delta &= P(Y_0=1, Y_1=1) \label{eq:delta}
\end{align}
From which we can derive
\begin{align}
    S_0&=P(Y_0=1)=P(Y_0=1,Y_1=0)+P(Y_0=1,Y_1=1)=\beta+\delta\label{eq:id_S_0}\\
    S_1&=P(Y_1=1)=P(Y_0=0,Y_1=1)+P(Y_0=1,Y_1=1)=\gamma+\delta.\label{eq:id_S_1}
\end{align}

Note that the \emph{probability of necessity and sufficiency} (PNS) in~\citep{pearl2009causality} is the  $\gamma$ term in~\eqref{eq:gamma}.

In customer churn prevention, the four counterfactuals may be mapped to the four categories of customers presented in the introduction (Table~\ref{tab:customer_categories}). An effective campaign should then only reach out  to \emph{persuadable} customers (whose proportion in the  population is $\beta$), since \emph{sure-thing} and \emph{lost cause} customers would not change their minds in reaction to the marketing action, and the \emph{do-not-disturb} would react negatively to it. 

The next sections will discuss the paper's contributions on the estimation of the probabilities $\alpha,\beta,\gamma$ and $\delta$.

\begin{table}
    \begin{center}
    \begin{minipage}{0.55\textwidth}
    \caption{The four categories of customers for churn prevention in terms of counterfactual outcomes.}\label{tab:customer_categories}%
    \begin{tabular}{c|cc}
        & $Y_0=0$ & $Y_0=1$ \\[1pt]
        \hline
        \rule{0pt}{1\normalbaselineskip}
        $Y_1=0$ & Sure thing & Persuadable\\
        $Y_1=1$ & Do-not-disturb & Lost cause
    \end{tabular}
    \end{minipage}
    \end{center}
\end{table}

\section{Bounds on the probability of counterfactuals}
\label{sec:bounds_cf}
Bounds on the probability of counterfactuals have first been derived in \citet{tian2000probabilities}, where the authors focus on $P(Y_0=0\mid T=1,Y=1)$, $P(Y_1=1\mid T=0,Y=0)$, and $P(Y_0=0,Y_1=1)$ (denoted $\gamma$ in~\eqref{eq:gamma}) under various assumptions. They showed that the quantity $\gamma$ can be bounded as
\begin{equation}
\label{eq:Pearlbound}
    \max\{0, P(Y_1=1) - P(Y_0=1)\}\le\gamma\le\min\{P(Y_0=0), P(Y_1=1)\}.
\end{equation} 
The bounds derive from the classical Fréchet bounds~\citep{frechet1935generalisation} stating 
that for any pair of events $A$ and $B$
\begin{equation}
    \label{eq:frechet}
    \max\{0, P(A)+P(B)-1\}\le P(A,B)\le\min\{P(A),P(B)\}.
\end{equation}
For instance, by replacing $A$ with $Y_0=0$ and $B$ with $Y_1=1$, it is easy to derive the inequalities~\eqref{eq:Pearlbound}. Tighter bounds on counterfactual probabilities  are derived in~\citep{mueller2021causes,zhang2022partial} by making structural assumptions on the causal directed acyclic graph (DAG). 

In this paper, we focus on a setting where (i) no structural assumptions may be made (besides unconfoundedness) and (ii) an estimation of the uplift 
is possible on the basis of historical data.
For this reason, we derive a set of original bounds that depend on the conditional probabilities terms $S_0(x)=P(Y_0=1\mid X=x)$ and $S_1(x)=P(Y_1=1\mid X=x)$.

Our derivation consists in first generalizing the Fréchet bounds to all four counterfactual probabilities, by substituting $A$ with $Y_0=0$ or $Y_0=1$, and $B$ with $Y_1=0$ or $Y_1=1$:
\begin{align}
    \max\{0, P(Y_0=0) - P(Y_1=1)\} &\le  \alpha \le \min\{P(Y_0=0), P(Y_1=0)\} \label{eq:frechet_a}\\
    \max\{0, P(Y_0=1) - P(Y_1=1)\}     &\le  \beta  \le \min\{P(Y_0=1), P(Y_1=0)\}\label{eq:frechet_b}\\
    \max\{0, P(Y_1=1) - P(Y_0=1)\}     &\le  \gamma \le \min\{P(Y_0=0), P(Y_1=1)\}\label{eq:frechet_c}\\
    \max\{0, P(Y_0=1) - P(Y_1=0)\} &\le  \delta \le \min\{P(Y_0=1), P(Y_1=1)\}.\label{eq:frechet_d}
\end{align}
Then, we assume that a reliable estimate (e.g. by uplift modeling) of the conditional scores $S_0(x)$ and $S_1(x)$ is available. 
Such scores can be used to refine the bounds on $\alpha,\dots,\delta$ by leveraging Jensen's inequality\footnote{Jensen's inequality, in its probabilistic form, states that for a convex function $f$ and a random variable $X$, we have
\[f(\E[X])\le \E[f(X)].\]}.
We apply first Jensen's inequality to the lower bounds of Equations~\eqref{eq:frechet_a}-\eqref{eq:frechet_d} by taking $f$ as the $\max(0,\cdot)$ function and
then to the upper bounds with $f$ as the $\min(\cdot,\cdot)$ function. 
We detail here the derivation for the lower bound on $\beta$, but the same reasoning can be easily extended   to the other bounds as well.
\begin{align}
    \max\{0, P(Y_0=1)-P(Y_1=1)\}&=\max\{0, S_0 - S_1\} \label{eq:deriv_jensen_1}\\
    &=\max\{0, \E[S_0(X) - S_1(X)]\} \\
    &\le \E[\max\{0, S_0(X) - S_1(X)\}] \label{eq:deriv_jensen_2}\\
    &\le \E[\beta(X)] = \beta.\label{eq:deriv_jensen_3}
\end{align}
Note that the quantity in~\eqref{eq:deriv_jensen_1} is the Fréchet bound, which by Jensen's inequality is lower than~\eqref{eq:deriv_jensen_2}. It follows that our  derivation returns a tighter upper bound than the Fréchet upper bound. The inequality~\eqref{eq:deriv_jensen_3}, derived from~\eqref{eq:frechet_b} conditioned on $X=x$, guarantees that this is a lower bound on $\beta$. To summarize, we propose to bound $\alpha,\dots,\delta$ as follows
\begin{align}
    \E[\max\{0, 1 - S_0(X) - S_1(X)\}] &\le  \alpha \le \E[\min\{1 - S_0(X), 1 - S_1(X)\}]\label{eq:bounds_a}\\
    \E[\max\{0, S_0(X) - S_1(X)\}]     &\le  \beta  \le \E[\min\{S_0(X), 1 - S_1(X)\}]\label{eq:bounds_b}\\
    \E[\max\{0, S_1(X) - S_0(X)\}]     &\le  \gamma \le \E[\min\{1 - S_0(X), S_1(X)\}]\label{eq:bounds_c}\\
    \E[\max\{0, S_0(X) + S_1(X) - 1\}] &\le  \delta \le \E[\min\{S_0(X), S_1(X)\}].\label{eq:bounds_d}
\end{align}
Hereafter we will refer to those bounds as the  \emph{uplift bounds} (UB) since they are defined in terms of the uplift terms. To assess whether
 these bounds improve the state-of-the-art Fréchet bounds, we consider their respective spans (i.e. the difference between the upper and the lower bound). It can be shown that the uplift bounds span $\Span_{\mathrm{UB}}$ is the same for all the counterfactual probabilities:
\begin{align}
    \Span_{\mathrm{UB}} &= \E[\min\{S_0(X), 1 - S_1(X)\}] - \E[\max\{0, S_0(X) - S_1(X)\}] \\
    &= \E[\min\{S_0(X), 1 - S_1(X)\} - \max\{0, S_0(X) - S_1(X)\}] \\
    &= \E[\min\{S_0(X), 1 - S_1(X)\} + \min\{0, S_1(X) - S_0(X)\}] \label{eq:deriv1}\\
    &= \E[\min\{S_0(X), S_1(X), 1 - S_0(X), 1-S_1(X)\}] \label{eq:deriv2}
\end{align}
Where in~\eqref{eq:deriv1} we used the equality $-\max\{a,b\}=\min\{-a,-b\}$, and in~\eqref{eq:deriv2} the equality $\min\{a,b\}+\min\{c,d\}=\min\{a+c,a+d,b+c,b+d\}$.

The span of the Fréchet bounds, denoted by $\Span_{\mathrm{Fr}}$, is equal to \[\Span_{\mathrm{Fr}} = \min\{S_0, S_1, 1-S_0, 1-S_1\}\] For all four counterfactual probabilities. 
Note that $\Span_{\mathrm{Fr}}$ depends solely on the marginal terms $S_0$ and $S_1$ (i.e. the average probability of the outcome in the control and target groups) whereas $\Span_{\mathrm{UB}}$ is a function of the descriptive features (or covariates) $X$. This means that in the case of informative features (i.e. when the conditional entropy of $Y_0$ and $Y_1$ is smaller than the marginal entropy), the uplift bounds are tighter than the Fréchet ones.
In the case of perfect knowledge (i.e. when $Y_0$ and $Y_1$ are deterministic functions of $X$), $S_0(x)$ and $S_1(x)$ are either $0$ or $1$, the span of the uplift bounds collapses to zero and the counterfactual distribution is fully determined. In the case 
of noninformative features (i.e. when the conditional entropy of $Y_0$ and $Y_1$ is equal to the marginal entropy) the uplift bounds reduce to the Fréchet bounds.

Such considerations can be formalized in terms of conditional entropy by the following Theorem:

\begin{theorem}
    \label{thm:bounds_span}
    As the conditional entropy $H(Y_0,Y_1\mid X)$ approaches zero, the uplift bounds on the probability $P(Y_0=y_0,Y_1=y_1)$ collapse to the exact value of that probability. Conversely, as the conditional entropy $H(Y_0,Y_1\mid X)$ approaches the entropy $H(Y_0,Y_1)$, the uplift bounds reduce to the Fréchet bounds.
    \begin{proof}
    We have
    \begin{equation*}
        H(Y_0,Y_1\mid X)=-\int\sum_{y_0,y_1} P(y_0,y_1\mid x)\log P(y_0,y_1\mid x) f_X(x)\dd x
    \end{equation*}
    Where the sum runs over the four possible values of $Y_0,Y_1$, and $f_X(x)$ is the probability density function of $X$. This expression can also be noted
    \begin{align*}
        H(Y_0,Y_1\mid X)=-\int(&\alpha(x)\log(\alpha(x))+\beta(x)\log(\beta(x))\\
        +&\gamma(x)\log(\gamma(x))+\delta(x)\log(\delta(x))) f_X(x)\dd x.
    \end{align*}
    It is minimized (in fact, equal to zero) when one of $\alpha(x),\dots,\delta(x)$ is equal to one and the three other ones are equal to zero for all $x$. Also, the span of the uplift bounds is
    \begin{align*}
        \Span_{\mathrm{UB}} &= \E[\min\{S_0(X),S_1(X),1-S_0(X),1-S_1(X)\}] \\
        &= \int \min\{\beta(x)+\delta(x),\gamma(x)+\delta(x),\alpha(x)+\gamma(x),\alpha(x)+\beta(x)\}f_X(x)\dd x
    \end{align*}
    When one of $\alpha(x),\dots,\delta(x)$ is equal to one and the three other values are equal to zero for all $x$, this expression collapses to zero, since two of the four terms in the minimum will be equal to zero. In this case, the bounds collapse to the true value of the counterfactual probability. This proves the first part of the theorem.

    For the second part of the theorem, let's assume that $X$ brings no information about $Y_0,Y_1$, which we formalize as $H(Y_0,Y_1\mid X)=H(Y_0,Y_1)$, or also in terms of statistical independence as $(Y_0,Y_1)\perp X$. By definition of statistical independence, we know that $P(y_0\mid x)=P(y_0)$ and $P(y_1\mid x)=P(y_1)$ for all values $y_0,y_1$ and $x$. Hence, as an example for $\beta$, the uplift bounds simplify to
    \begin{align*}
        \E[\max\{0,P(Y_0=1)-P(Y_1=1)\}]&\le\beta\le\E[\min\{P(Y_0=1),P(Y_1=1)\}].
    \end{align*}
    The expected value is on the distribution of $X$, but since the terms in the expected value do not depend on $X$, the bounds reduce to
    \begin{align*}
        \max\{0,S_0-S_1\}&\le\beta\le\min\{S_0,S_1\}
    \end{align*}
    Which are the Fréchet bounds on $\beta$. The same reasoning applies to the bounds on $\alpha,\gamma$ and $\delta$.
    \end{proof}
\end{theorem}

\subsection{Probability bounds and uplift estimation}
The  main motivation underlying the derivation of the uplift bounds is that in real-world settings characterized by large historical data sets (like churn modeling), it is possible to derive sample-based estimates of the terms bounding the counterfactual probabilities.
In particular, we advocate the adoption of a plug-in estimator from an uplift model $\widehat S_0(x), \widehat S_1(x)$ on a data set $\mathcal D=\{x^{(1)},\dots,x^{(N)}\}$. In this case, a sample-based version of  the lower bound on $\beta$ is
\begin{equation}
    \E[\max\{0, S_0(X) - S_1(X)\}] \approx \frac{1}{N}\sum_{i=1}^N \max\left\{0, \widehat S_0\left(x^{(i)}\right) - \widehat S_1\left(x^{(i)}\right)\right\}
\end{equation}
And similarly for the other bounds on $\alpha,\dots,\delta$.

It is sometimes desirable to obtain a point estimate on the probability of counterfactuals, for example when a unique number is expected as the result of the counterfactual analysis. Though a naive estimator could be derived by taking the midpoint of the bounds, in the next section we will introduce a more theoretically founded estimator.

\section{Point estimate of counterfactual probabilities}
\label{sec:point_cf}
Counterfactual probabilities are latent yet very important quantities to be taken into consideration for decision-making. In the previous section, we proposed an original approach to bound their values. However, it is sometimes desirable to compute a point estimate of those probabilities, even if this requires stronger assumptions. Here we present a point estimator of the  probabilities $\alpha,\dots,\delta$ (Equations~\eqref{eq:alpha} to~\eqref{eq:delta}) based on the conditional independence  between $Y_0$ and $Y_1$. The introduction of specific assumptions is required
since those probabilities, e.g.
\begin{equation}
    \label{eq:point_derivation_1}
    \alpha=P(Y_0=0,Y_1=0)=\E_X[P(Y_0=0,Y_1=0\mid X)]=\E_X[\alpha(X)]
\end{equation}
Cannot be estimated from observational or experimental data, given that one of the two outcomes will be necessarily unobserved. The conditional independence between $Y_0$ and $Y_1$ given $X=x$, which is formally expressed as $Y_0\perp Y_1\mid X=x$, allows developing the term $\alpha(x)$ in Equation~\eqref{eq:point_derivation_1} as
\begin{align}
    \alpha(x)&\approx P(Y_0=0\mid X=x)P(Y_1=0\mid X=x).
\end{align}
In order to study the impact of the 
conditional independence assumption, we define the difference between $P(Y_0=0,Y_1=0\mid X=x)$ and the approximation $P(Y_0=0\mid X=x)P(Y_1=0\mid X=x)$ as $\phi(x)$. This quantity appears in the other conditional probabilities too:
\begin{align}
    \alpha(x) &= P(Y_0=0\mid X=x)P(Y_1=0\mid X=x) + \phi(x) \label{eq:alpha_x} \\
    \beta(x)  &= P(Y_0=1\mid X=x)P(Y_1=0\mid X=x) - \phi(x) \label{eq:beta_x} \\
    \gamma(x) &= P(Y_0=0\mid X=x)P(Y_1=1\mid X=x) - \phi(x) \label{eq:gamma_x}\\
    \delta(x) &= P(Y_0=1\mid X=x)P(Y_1=1\mid X=x) + \phi(x).\label{eq:delta_x}
\end{align}

Note that the quantity $\phi(x)$ can be interpreted as a conditional measure of dependency between $Y_0$ and $Y_1$ and
is similar to classical binary dependency measures, like the odd ratio, Yule's $Q$ coefficient, or the difference coefficient~\citep{edwards1957note}. 
We will see in Theorem~\ref{thm:bias_ml_est} that
\begin{equation}
\phi=\alpha\delta-\beta\gamma-\mathrm{cov}_X(S_0(X),S_1(X)).
\end{equation}
Meaning that $\phi$ depends both on the distribution of counterfactuals ($\alpha,\beta,\gamma$ and $\delta$) and the dependency between the scores $S_0(x)$ and $S_1(x)$.
From~\eqref{eq:point_derivation_1} we obtain
\begin{align}
    \alpha &= \E[\alpha(X)] \\
          &= \E[P(Y_0=0\mid X)P(Y_1=0\mid X)+\phi(X)] \\
          &= \E[(1-S_0(X))(1-S_1(X))]+\phi\label{eq:point_derivation_2}
\end{align}
Where $\phi=\E[\phi(X)]$. If we  assume $Y_0\perp Y_1\mid X$ ,
then $\phi=0$ and
\begin{equation}
    \alpha\approx\E[(1-S_0(X))(1-S_1(X))].
\end{equation}

The question of the dependency between $Y_0$ and $Y_1$ has already been discussed in the causal inference literature~\citep[Sec. 8.6]{imbens2015causal}. A possible approach could be to assume the maximum possible dependency between the potential outcomes. Alternatively, one could make no a priori preference between a positive and negative association between $Y_0$ and $Y_1$ (i.e. $Y_0$ and $Y_1$ taking similar or opposite values), thus assuming no association. 
Since, in absence of some preexisting knowledge, there is no a priori good answer, it is more interesting to  reason about the dependency between $Y_0$ and $Y_1$ as follows:
\begin{itemize}
    \item A positive correlation\footnote{The correlation between $Y_0$ and $Y_1$ refers to the tendency of $Y_0$ and $Y_1$ to take identical or opposite values.} between $Y_0$ and $Y_1$ means that they are often equal, indicating that the treatment has little effect on the outcome. When the correlation is maximum, the upper bounds on $\alpha$ and $\delta$ in Equations~\eqref{eq:bounds_a} and \eqref{eq:bounds_d} are met.
    \item A negative correlation between $Y_0$ and $Y_1$ indicates that the treatment has either a strongly positive or negative impact on the outcome.  When the correlation is maximally negative, the upper bounds on $\beta$ and $\gamma$ in Equations~\eqref{eq:bounds_b} and \eqref{eq:bounds_c} are met.
    \item The absence of dependency indicates an even mix of the two previous cases. This corresponds to the point estimator presented in this section.
\end{itemize}

\subsection{Point estimate and uplift estimation}
Given estimators $\widehat S_0(x),\widehat S_1(x)$ of the uplift terms, and an evaluation data set $\{x^{(i)}\}_{i=1,\dots,N}$, we propose to estimate $\alpha,\dots,\delta$ as
\begin{align}
    \hat\alpha &= \frac{1}{N}\sum_i (1-\widehat S_0(x^{(i)})(1-\widehat S_1(x^{(i)}))\label{eq:alpha_hat}\\
    \hat\beta &= \frac{1}{N}\sum_i \widehat S_0(x^{(i)})(1-\widehat S_1(x^{(i)}))\label{eq:beta_hat}\\
    \hat\gamma &= \frac{1}{N}\sum_i (1-\widehat S_0(x^{(i)}))\widehat S_1(x^{(i)})\label{eq:gamma_hat}\\
    \hat\delta &= \frac{1}{N}\sum_i \widehat S_0(x^{(i)})\widehat S_1(x^{(i)}).\label{eq:delta_hat}
\end{align}
The bias of these estimators is expressed in Theorem~\ref{thm:bias_ml_est}.

\begin{theorem}
    \label{thm:bias_ml_est}
    Given that $\widehat S_0(x)$ and $\widehat S_1(x)$ are unconfounded and unbiased estimators of $S_0(x)$ and $S_1(x)$ trained on a training set with distribution $D$, in the large sample limit the bias of $\hat\alpha,\dots,\hat\delta$ is
    \begin{align}
        \mathrm{Bias}[\hat\beta]&=\mathrm{Bias}[\hat\gamma]=-\mathrm{Bias}[\hat\alpha]=-\mathrm{Bias}[\hat\delta]\nonumber\\
        &=\alpha\delta-\beta\gamma-\mathrm{cov}_X(S_0(X),S_1(X))-\E_X[\mathrm{cov}_{D}(\widehat S_0(X),\widehat S_1(X))]\label{eq:bias_ml_est_1}\\
        &=\phi-\E_X[\mathrm{cov}_{D}(\widehat S_0(X),\widehat S_1(X))].\label{eq:bias_ml_est_2}
    \end{align}
    \begin{proof}
    We will derive the bias of $\hat\beta$, and the bias of the three other estimators can be derived in a similar way. The expected value of $\hat\beta$ over the distribution of training sets $D$ is
        \begin{align*}
            \E_{D}[\hat\beta]&=\E_{D}\left[\frac{1}{N}\sum_{i=1}^N\widehat S_0(x^{(i)})(1-\widehat S_1(x^{(i)}))\right]\\
            &=\frac{1}{N}\sum_{i=1}^N\E_{D}\left[\widehat S_0(x^{(i)})(1-\widehat S_1(x^{(i)}))\right]\\
            &=\frac{1}{N}\sum_{i=1}^N\E_{D}[\widehat S_0(x^{(i)})]\E_{D}[1-\widehat S_1(x^{(i)})]+\mathrm{cov}_{D}(\widehat S_0(x^{(i)}),1-\widehat S_1(x^{(i)}))\\
            &=\frac{1}{N}\sum_{i=1}^NS_0(x^{(i)})(1-S_1(x^{(i)}))-\mathrm{cov}_{D}(\widehat S_0(x^{(i)}),\widehat S_1(x^{(i)})).
        \end{align*}
        In the large sample limit ($N\rightarrow +\infty$), we can assume that this sum converges to
        \begin{align*}
        \E_{D}[\hat\beta]&=\E_X[S_0(X)(1-S_1(X))]-\E_X[\mathrm{cov}_{D}(\widehat S_0(X),\widehat S_1(X)].
        \end{align*}
        The first term can be expanded as
        \begin{align*}
            \E[S_0(X)(1-S_1(X))]&=\E[S_0(X)]\E[1-S_1(X)]+\mathrm{cov}_X(S_0(X),1-S_1(X))\\
            &=S_0(1-S_1)-\mathrm{cov}_X(S_0(X),S_1(X))\\
            &=(\beta+\delta)(\beta+\alpha)-\mathrm{cov}_X(S_0(X),S_1(X))\\
            &=\beta(\beta+\delta+\alpha)+\alpha\delta-\mathrm{cov}_X(S_0(X),S_1(X))\\
            &=\beta(1-\gamma)+\alpha\delta-\mathrm{cov}_X(S_0(X),S_1(X))\\
            &=\alpha\delta-\beta\gamma+\beta-\mathrm{cov}_X(S_0(X),S_1(X)).
        \end{align*}
        And thus
        \begin{align*}
        \E_{D}[\hat\beta]&=\alpha\delta-\beta\gamma+\beta-\mathrm{cov}_X(S_0(X),S_1(X))-\E_X[\mathrm{cov}_{D}(\widehat S_0(X),\widehat S_1(X)].
        \end{align*}
        Finally, the bias of $\hat\beta$ is
        \begin{align*}
             \mathrm{Bias}[\hat\beta]&=\E_{D}[\hat\beta]-\beta\\
             &=\alpha\delta-\beta\gamma-\mathrm{cov}_X(S_0(X),S_1(X))-\E_X[\mathrm{cov}_{D}(\widehat S_0(X),\widehat S_1(X)]
        \end{align*}
        Which proves Equation~\eqref{eq:bias_ml_est_1}. Equation~\eqref{eq:bias_ml_est_2} is derived from  \begin{align*}
            \E[S_0(X)(1-S_1(X))]&=\E[\beta(x)+\phi(x)]
            =\beta+\phi
        \end{align*}
        And then
        \begin{align*}
            \mathrm{Bias}[\hat\beta]&=\E_{D}[\hat\beta]-\beta\\
            &=\E_X[S_0(X)(1-S_1(X))]-\E_X[\mathrm{cov}_{D}(\widehat S_0(X),\widehat S_1(X)]-\beta\\
            &=\phi-\E_X[\mathrm{cov}_{D}(\widehat S_0(X),\widehat S_1(X)].
        \end{align*}
    \end{proof}
\end{theorem}
While the three first terms in Equation~\eqref{eq:bias_ml_est_1} are inherent to the customer population, the last term depends also on the estimators $\widehat S_0(x)$ and $\widehat S_1(x)$, and the data distribution $D$. Without assumptions about these processes, the last term cannot be further reduced.

The proposed procedure to compute $\hat\beta$ as well as the two uplift bounds on $\beta$ presented in Section~\ref{sec:bounds_cf} is described in Algorithm~\ref{alg:estimation}, where we assume we have two unbiased estimators of the scores $S_0(x)$ and $S_1(x)$.
\begin{algorithm}
   \caption{Estimating the counterfactual probability $\beta$}
    \label{alg:estimation}
    \begin{algorithmic}
        \State {\bfseries Input:} Data set $\mathcal D=\{(x^{(i)}, y^{(i)}, t^{(i)})\}_{i=1,\dots,N}$
        \State {\bfseries Output:} Point estimate $\hat\beta$, and bounds $\widehat{\mathrm{LB}}_\beta$ and $\widehat{\mathrm{UB}}_\beta$ such that $\widehat{\mathrm{LB}}_\beta\le\beta\le\widehat{\mathrm{UB}}_\beta$
        \State Split $\mathcal D$ into training set $\mathcal D_{tr}$ and test set $\mathcal D_{te}$
        \State Train uplift model on $\mathcal D_{tr}$ to obtain estimators $(\widehat S_0(x), \widehat S_1(x))$
        \State Compute $\hat\beta = \frac{1}{\vert\mathcal{D}_{te}\vert}\sum_i \widehat S_0(x^{(i)})(1-\widehat S_1(x^{(i)}))$ on $\mathcal D_{te}$
        \State Compute $\widehat{\mathrm{LB}}_\beta = \frac{1}{\vert\mathcal D_{te}\vert}\sum_i \max\{0, \widehat S_0(x^{(i)}) - \widehat S_1(x^{(i)})\}$ on $\mathcal D_{te}$
        \State Compute $\widehat{\mathrm{UB}}_\beta = \frac{1}{\vert\mathcal D_{te}\vert}\sum_i \min\{\widehat S_0(x^{(i)}), 1 - \widehat S_1(x^{(i)})\}$ on $\mathcal D_{te}$
    \end{algorithmic}
\end{algorithm}

\section{Bounds assessment by simulation}
\label{sec:simulated}
In this section, we assess the bounds and estimators presented in Sections~\ref{sec:bounds_cf} and \ref{sec:point_cf} by setting up a specific simulation environment. The simulated nature of the
experiment allows us to compare the estimated bounds to the ground truth.

\subsection{Methodology}
Let $\alpha,\beta,\gamma$ and $\delta$ be the terms introduced
in~\eqref{eq:alpha}, \eqref{eq:beta}, \eqref{eq:gamma}, \eqref{eq:delta}. 
The aim of the simulation is to generate samples from 
a distribution where the  scores $S_0(x)$ and $S_1(x)$ are conditional on a set of features $X$.

One possible approach is to model the $X$ covariate distribution, the stochastic functional dependency between $Y$, $X$ and $T$, train an uplift model $\widehat S_0(x),\widehat S_1(x)$ on a generated data set $\mathcal D=\{(x^{(i)},y^{(i)},t^{(i)})_{i=1,\dots,N}\}$, and finally apply the estimators presented in the previous sections. We did not consider this approach since the results would heavily depend on the model choices (e.g. the distribution of $X$ and the class of functions for $Y$) and the learning algorithm. 

Our simulation setting consists in directly sampling the distributions
of the estimators $\widehat S_0$ and $\widehat S_1$ 
obtained as a noisy version of $S_0$ and $S_1$ 
which, according to~\eqref{eq:id_S_0} and~\eqref{eq:id_S_1}, are functions of the terms $\alpha,\dots,\delta$.
Since we sample the distribution of scores $\widehat S_0$ and $\widehat S_1$ but we do not sample $X$ directly, we will denote individual scores with superscript $i$ rather than as functions of $x$. The sampling process of our simulation is detailed in Equations~\eqref{eq:sample_mu} to \eqref{eq:sample_y}:
\begin{align}
    &(\alpha^{(i)},\dots,\delta^{(i)})\sim\Dir(a,b,c,d)\label{eq:sample_mu}\\
    &S_0^{(i)}=\beta^{(i)}+\delta^{(i)}    \quad\;\;\, S_1^{(i)} =\gamma^{(i)}+\delta^{(i)}\label{eq:sample_score}\\
    &\widehat S_0^{(i)}\sim\frac1v\Binom(v,S_0^{(i)}) \quad \widehat S_1^{(i)}\sim\frac1v\Binom(v,S_1^{(i)})\label{eq:sample_hat}\\
    &(Y_0^{(i)},Y_1^{(i)})\sim\Cat(\alpha^{(i)},\dots,\delta^{(i)})\label{eq:sample_y}
\end{align}

First, we generate $N$ independent samples $(\alpha^{(i)}, \beta^{(i)}, \gamma^{(i)},\delta^{(i)})_{i=1,\dots, N}$ according to a Dirichlet distribution $\Dir(a,b,c,d)$. They represent the probabilities of counterfactuals at the individual level. The Dirichlet distribution is a natural candidate to sample numbers in a probability simplex (i.e. such that $\alpha^{(i)}, \beta^{(i)}, \gamma^{(i)}$ and $\delta^{(i)}$ are all positive and sum up to $1$), since it is the conjugate prior of the multinomial distribution~\citep{lin2016dirichlet}. Then, we derive the value of the scores $S_0^{(i)}$ and $S_1^{(i)}$ with the identities $S_0^{(i)}=\beta^{(i)}+\delta^{(i)}$ (Equation~\eqref{eq:id_S_0}) and $S_1^{(i)}=\gamma^{(i)}+\delta^{(i)}$ (Equation~\eqref{eq:id_S_1}). To emulate imperfect estimators $\widehat S_0^{(i)}$ and $\widehat S_1^{(i)}$, we draw $\widehat S_t^{(i)}$ (for $t=0,1$) according to a normalized binomial distribution $\frac{1}{v}\mathrm B(v, S_t^{(i)})$, where $v$ is the parameter controlling the variance of $\widehat S_t^{(i)}$. Such estimator distribution
guarantees that $\widehat S_t^{(i)}$ takes  values inside $[0, 1]$ and models 
the variability of $\widehat S_t^{(i)}$ due to a limited number of training examples of a binary outcome $Y_t$. Finally, the counterfactual outcomes $Y_0^{(i)}$ and $Y_1^{(i)}$ are sampled according to a categorical distribution $\Cat(\alpha^{(i)},\dots,\delta^{(i)})$ such that $P(Y_0^{(i)}=0,Y_1^{(i)}=0)=\alpha^{(i)}$, and similarly for $\beta^{(i)}$, $\gamma^{(i)}$ and $\delta^{(i)}$, reflecting Equations~\eqref{eq:alpha} to~\eqref{eq:delta}. Once the sampling process is executed, the bounds and estimators from Sections~\ref{sec:bounds_cf} and \ref{sec:point_cf} can be evaluated from the set of scores $\{(\widehat S_0^{(i)},\widehat S_1^{(i)})\}_{i=1,\dots,N}$.

\subsection{Simulation parameters}
The simulation setting is defined by six main parameters: $N,v,a,b,c$ and $d$. 
\begin{itemize}
    \item The parameter $N$ represents the size of the data set on which the bounds and estimators are evaluated.
    \item The parameter $v$ emulates the variance of the simulated uplift model. Higher values of $v$ induce a lower variance since we can show\footnote{The variable $\widehat S_t^{(i)}$ is based on a Binomial distribution $\Binom(v,S_t^{(i)})$, which has a variance $vS_t^{(i)}(1-S_t^{(i)})$. We can develop the variance $\Var(\widehat S_t^{(i)})=vS_t^{(i)}(1-S_t^{(i)})/v^2=S_t^{(i)}(1-S_t^{(i)})/v$.} that $\Var(\widehat S_t^{(i)})=S_t^{(i)}(1-S_t^{(i)})/v$.
    \item The parameters $a,b,c$ and $d$ are proportional to the distribution of counterfactuals $P(Y_0^{(i)}=0,Y_1^{(i)}=1),\dots,P(Y_0^{(i)}=1,Y_1^{(i)}=1)$. For example, using the moments of the Dirichlet distribution, we have
\begin{equation}
    P(Y_0^{(i)}=1,Y_1^{(i)}=0)=\mathbb E[\beta^{(i)}] = \frac{b}{A}
    \label{eq:sim_beta}
\end{equation}
Where $A=a+b+c+d$.
    \item The value of $A$ influences the distribution of $\alpha^{(i)}, \dots,\delta^{(i)}$. High values of $A$ lead to samples $\alpha^{(i)}, \dots,\delta^{(i)}$ concentrated around their expected values (which can be computed from Equation~\eqref{eq:sim_beta}), while low values of $A$ lead to samples where one of $\alpha^{(i)}, \dots,\delta^{(i)}$ is close to one while the three other values are close to zero. This has an impact on the scores $S_0^{(i)},S_1^{(i)}$ as well: they are close to their expected values when $A$ is large, and close to either zero or one when $A$ is low. In loose terms, the quantity $A$
    represents the amount of information that the  covariates $X$ brings about the outcomes $Y_0$ and $Y_1$: when the features are uninformative, the scores $S_0(x),S_1(x)$ are close to their prior probabilities $P(Y_0=1)$ and $P(Y_1=1)$, while when the features are highly informative, the scores are close to either zero or one.
\end{itemize}

Theorem~\ref{thm:bias_ml_est} indicates that the bias of the points estimators $\hat\alpha,\dots,\hat\delta$ (transposed to the notation of this section) is
\[\E[\phi^{(i)}]-\E_{\alpha^{(i)},\dots,\delta^{(i)}}[\mathrm{cov}(\widehat S_0^{(i)},\widehat S_1^{(i)})].\]
The second term is null because we sample $\widehat S_0^{(i)}$ and $\widehat S_1^{(i)}$ independently, but we can show using the product moments of the Dirichlet distribution~\citep{lin2016dirichlet} that the first term $\E[\phi^{(i)}]$ is
\begin{equation}
\label{eq:bias_simulation}
\mathbb E[\phi^{(i)}]=\mathbb E[\alpha^{(i)}\delta^{(i)}-\beta^{(i)}\gamma^{(i)}]=\frac{ad-bc}{A(A+1)}.
\end{equation}
Since the parameters $a,b,c,d$ are sampled uniformly, the expression in Equation~\eqref{eq:bias_simulation} will be different from zero. Therefore, the distribution of the bias of the point estimators $\hat\alpha,\dots,\hat\delta$ has a large variance. This is desirable to assess how violations of the hypothesis underlying our estimators affect the quality of the estimation.

\subsection{Assessment of the theoretical results}
In this section, we assess the quality of the uplift bounds and the point estimator, discussed in Sections~\ref{sec:bounds_cf} and ~\ref{sec:point_cf} respectively,  for different values of the simulation parameters. The simulation process is repeated 5000 times with randomly chosen parameters. The size $N$ of the evaluation set varies between $10$ and $10000$ and the variance parameter $v$ varies between 5 and 50. The Dirichlet parameters $a,b,c,d$ are fixed as $(a,b,c,d)=A(\alpha,\beta,\gamma,\delta)$ where $A$ varies between $0.1$ and $15$, and the vector $(\alpha,\beta,\gamma,\delta)$ is sampled uniformly over the probability simplex (i.e. such that the four terms are positive and sum up to one).

\begin{figure}
    \vskip 0.2in
    \begin{center}
    \centerline{\includegraphics[width=0.8\columnwidth]{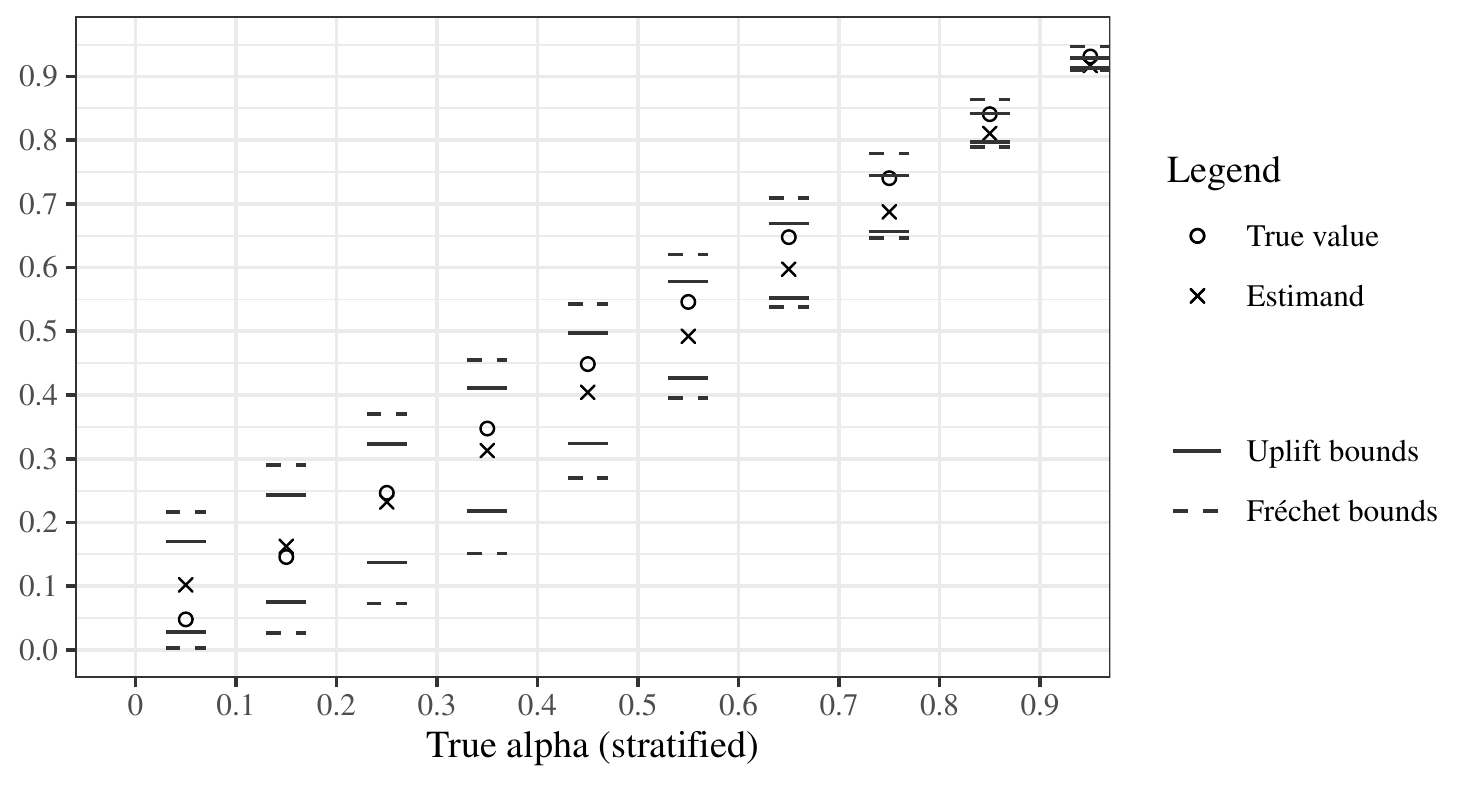}}
    \caption{The estimator $\hat\alpha$, the true value of $\alpha$, and the bounds on $\alpha$, for different values of $\alpha$. We take the average over all experiments where $\alpha$ falls into the relevant range. The graph is quite similar for $\beta,\gamma$ and $\delta$.}
    \label{fig:alpha_bounds}
    \end{center}
    \vskip -0.2in
\end{figure}

\begin{table}
    \begin{center}
    \begin{minipage}{0.8\textwidth}
    \centering
    \caption{Identification error for $\beta$. We compare the Fréchet bounds and the uplift bounds, and we also compare the point estimators with the bounds mid-point. We observe that the uplift bounds provide a clear improvement over the Fréchet bounds.}\label{tab:results_ml_pop}%
    \begin{tabular}{lc}
        \toprule
        \textbf{Bounds} & \textbf{Mean width} \\
        Uplift bounds       & 12.6\% \\
        Fréchet bounds  & 24.9\% \\
        \midrule
        \textbf{Estimator}       & \textbf{RMSE} \\
        Point estimator $\hat\beta$ & 6.4\% \\
        Uplift bounds mid-point         & 5.9\% \\
        Fréchet bounds mid-point    & 8.0\% \\
        \bottomrule
    \end{tabular}
    \end{minipage}
    \end{center}
\end{table}

\begin{figure}
    \vskip 0.2in
    \begin{center}
    \centerline{\includegraphics[width=0.7\columnwidth]{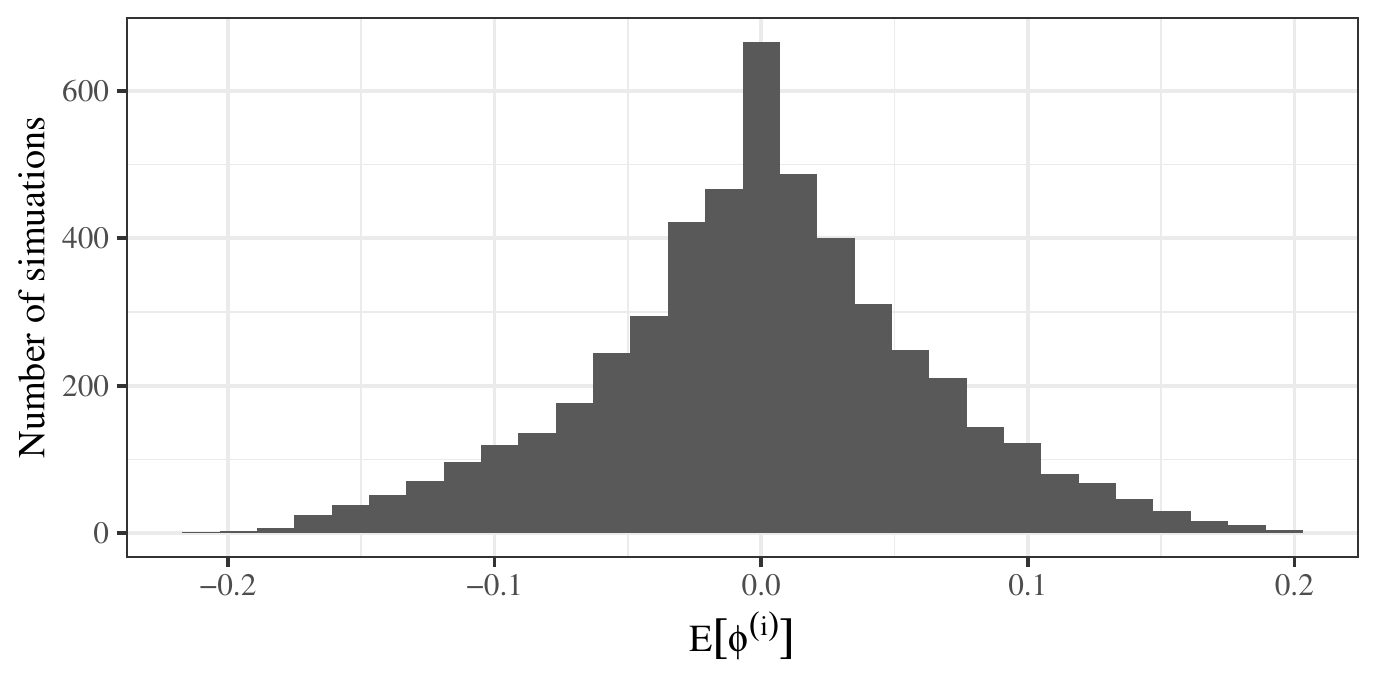}}
    \caption{Distribution of the point estimator bias, $\mathbb E[\phi^{(i)}]$, over 4000 simulation runs. Note that this is different from the distribution of $\phi^{(i)}$ in a given simulation run. Although the maximum is around zero, it is never exactly zero, indicating that the estimators are biased in our simulations. This is desirable to reflect violations of the hypotheses underlying our estimators in practical scenarios.}
    \label{fig:phi_distribution}
    \end{center}
    \vskip -0.2in
\end{figure}

Figure~\ref{fig:alpha_bounds} plots the estimator $\hat\alpha$ (cross), the uplift bounds (continuous line) and the Fréchet bounds (dashed lines) with respect to the true $\alpha$ (circle). Since the plots for $\beta,\gamma$ and $\delta$ are quite similar, they are omitted for the sake of conciseness. The values for the 5000 simulation runs are stratified according to the true value $\alpha$ in order to simplify the plot. For each stratum, the point reports the average of the estimated value and the horizontal bars report the average upper and lower bounds. 
The main conclusions of the simulation study are:

\begin{itemize}
    \item The uplift bounds are significantly tighter than the Fréchet bounds, as shown in Figure~\ref{fig:alpha_bounds}. The bounds span is typically reduced by half, as reported in Table~\ref{tab:results_ml_pop}.

    \item The point estimate provides a good approximation of the true counterfactual probability, with a root mean squared error (RMSE) of 6.4\% (Table~\ref{tab:results_ml_pop}). In order to have a baseline for comparison, we also compute the RMSE  of the bounds mid-point if those were taken as point estimators of the true counterfactual probability. We see that the Fréchet bounds mid-point has a larger error while the uplift bounds mid-point has an error comparable to the point estimate.

    \item The distribution of the bias $\mathbb E[\phi(x)]$ of the point estimator, which is defined by Equation~\eqref{eq:bias_simulation}, is  shown in Figure~\ref{fig:phi_distribution}. The fact that most of the bias realizations are different from zero is an indication of the realism of the simulation setting  and a positive sign about the robustness of the theoretical results.
  
\end{itemize}

\subsection{Sensitivity analysis of the simulation}
In this section, we assess the influence of the training data (in terms of the number of samples or the information of the features) on the precision of the estimation. We plot the span of the uplift bounds and the error of the point estimator while varying one of the parameters $A, N$ and $v$ and keeping the other parameters fixed. The values of the fixed parameters are selected to clearly show the influence of the varying parameters. In particular, we set $(\alpha,\beta,\gamma,\delta)=(0.947,0.020,0.017,0.017)$ based on the results of Section~\ref{sec:empirical}, which represents the distribution of counterfactuals in a typical scenario of customer churn prevention in telecom. The main conclusions of this sensitivity analysis are:

\begin{figure}
    \vskip 0.2in
    \begin{center}
    \centerline{\includegraphics[width=0.8\columnwidth]{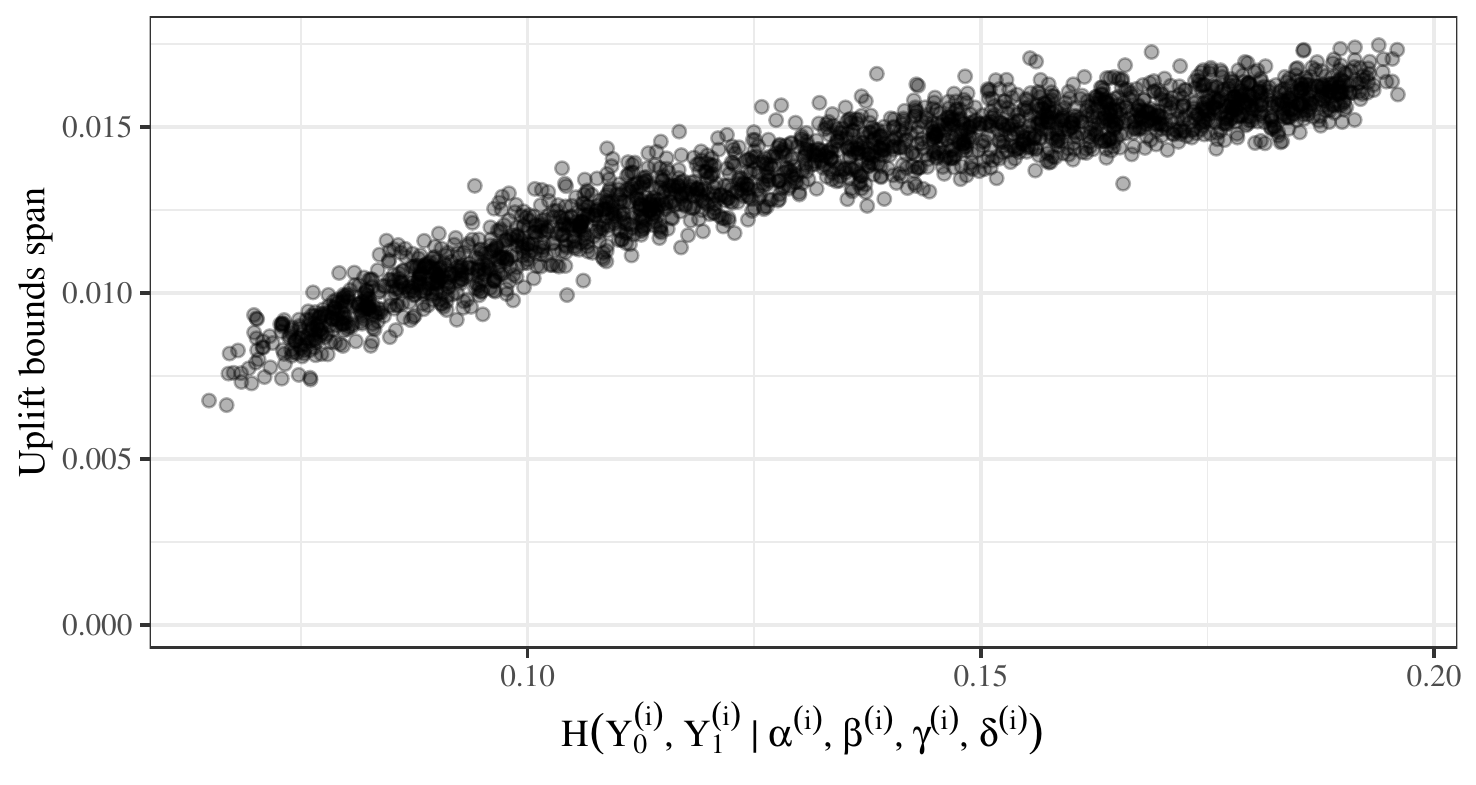}}
    \caption{The bounds span as a function of the conditional entropy of $Y_0^{(i)},Y_1^{(i)}$, which is directly influenced by the parameter $A$. We fixed $(\alpha,\beta,\gamma,\delta)=(0.947,0.020,0.017,0.017)$, and $v=50$ and $N=2000$.}
    \label{fig:entropy_bounds_span}
    \end{center}
    \vskip -0.2in
\end{figure}

\begin{figure}
    \vskip 0.2in
    \begin{center}
    \centerline{\includegraphics[width=0.8\columnwidth]{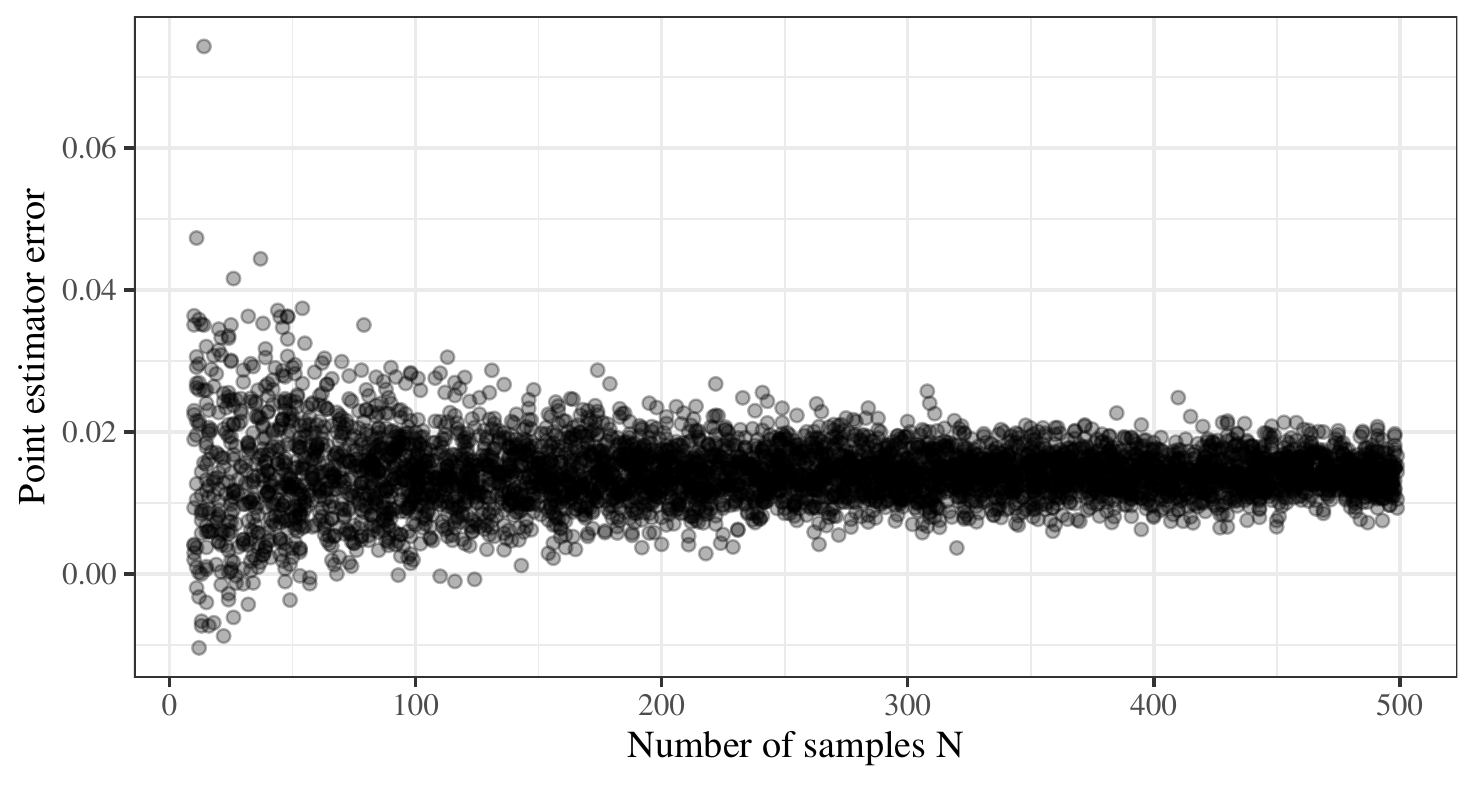}}
    \caption{The error of the point estimator as a function of the number of samples in the evaluation data set. We fixed $(\alpha,\beta,\gamma,\delta)=(0.947,0.020,0.017,0.017)$, and $v=20$ and $A=1$.}
    \label{fig:n_sample_error}
    \end{center}
    \vskip -0.2in
\end{figure}

\begin{figure}
    \vskip 0.2in
    \begin{center}
    \centerline{\includegraphics[width=0.8\columnwidth]{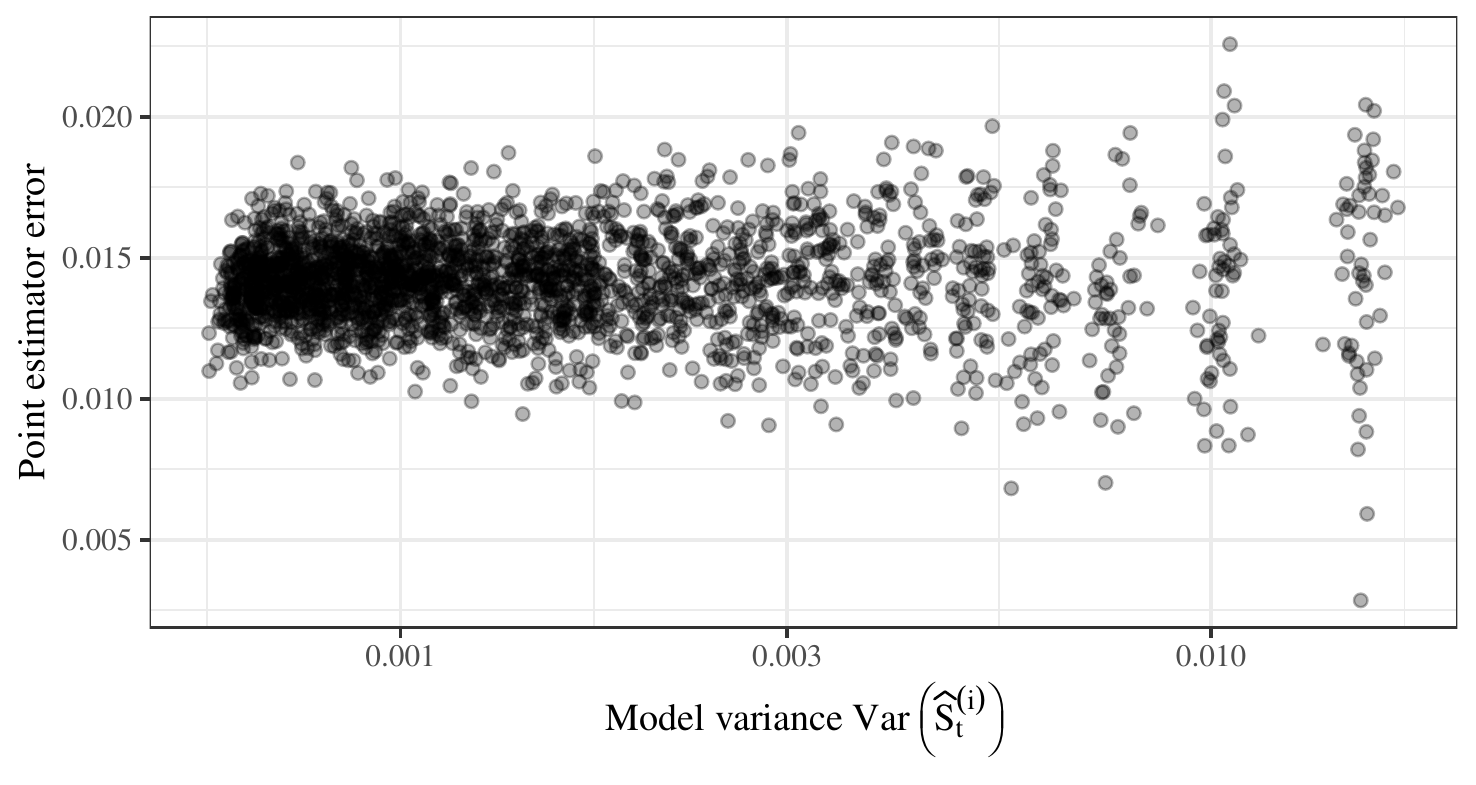}}
    \caption{The error of the point estimator as a function of model variance $\Var(\widehat S_t^{(i)})$. We fixed $(\alpha,\beta,\gamma,\delta)=(0.947,0.020,0.017,0.017)$, and $N=1000$ and $A=10$. As the variance decreases, the estimator bias converges towards to its theoretical value.}
    \label{fig:model_variance_error}
    \end{center}
    \vskip -0.2in
\end{figure}

\begin{figure}
    \vskip 0.2in
    \begin{center}
    \centerline{\includegraphics[width=0.8\columnwidth]{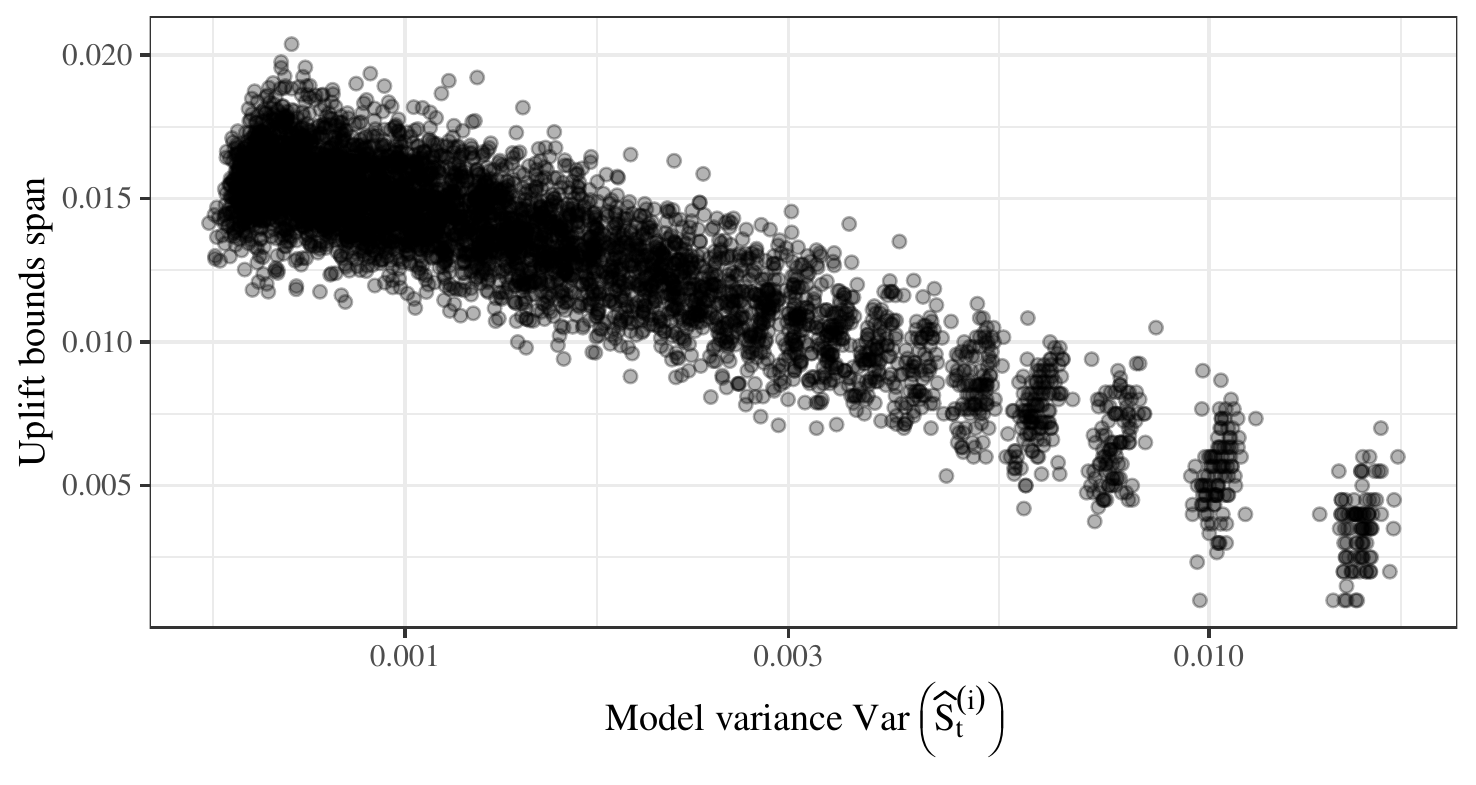}}
    \caption{The uplift bounds span as a function of the model variance $\Var(\widehat S_t^{(i)})$. We fixed $(\alpha,\beta,\gamma,\delta)=(0.947,0.020,0.017,0.017)$, and $N=1000$ and $A=10$. A lower model variance is shown here to be associated with larger bounds. In fact, as the variance goes to zero (left side of the plot), the bounds span converges towards its theoretical value. A model with a high variance predicts more often low values, which artificially reduces the bounds span.}
    \label{fig:model_variance_bounds_span}
    \end{center}
    \vskip -0.2in
\end{figure}

\begin{itemize}
    \item The uplift bounds span decreases as the conditional entropy of $Y_0,Y_1$ decreases (Figure~\ref{fig:entropy_bounds_span}). This is an empirical illustration of Theorem~\ref{thm:bounds_span}. In the context of this simulation, since we do not model the features $X$, we instead note the conditional entropy as $H(Y_0^{(i)},Y_1^{(i)}\mid\alpha^{(i)},\dots,\delta^{(i)})$, and we compute its value as
    \begin{align*}
        &H(Y_0^{(i)},Y_1^{(i)}\mid\alpha^{(i)},\dots,\delta^{(i)})\\
        &=\frac 1N \sum_{i=1}^N(-\alpha^{(i)}\log\alpha^{(i)}-\beta^{(i)}\log\beta^{(i)}-\gamma^{(i)}\log\gamma^{(i)}-\delta^{(i)}\log\delta^{(i)}).
    \end{align*}
    We see that as the conditional entropy approaches zero (which emulates very informative features), the bounds span converges towards zero as well.

    \item The variance of the point estimator decreases  as the number of samples $N$ increases (Figure~\ref{fig:n_sample_error}) or the model variance $\Var(\widehat S_t^{(i)})$ decreases (Figure~\ref{fig:model_variance_error}). In fact, the error converges towards the bias derived in Theorem~\ref{thm:bias_ml_est}. This demonstrates the convergence of our estimator in the large sample scenario.

    \item The uplift bounds span increases with the 
    decrease of the model variance $\Var(\widehat S_t^{(i)})$ (for $t=0,1$)  (Figure~\ref{fig:model_variance_bounds_span}). This is because a model with a high variance predicts often lower or higher scores than the expected score. Since the bounds span is 
    $\E[\min\{S_0^{(i)}, S_1^{(i)}, 1-S_0^{(i)}, 1-S_1^{(i)}\}]$ (see Equation~\eqref{eq:deriv2}), this artificially reduces the bounds span.
\end{itemize}

\section{Evaluation with real data}
\label{sec:empirical}

This section applies the theoretical results discussed so far
to a real-world data set provided by our industrial partner O.\footnote{The name of the company is hidden for the purpose of the blind review process.} that includes 6 churn prevention campaigns.

\subsection{Data set description}
Churn prevention campaigns are used to mitigate customer churn by contacting customers at risk of leaving the company. They are offered an incentive to stay, such as a promotional offer or a suggestion for a better tariff plan. The retention campaigns were performed over 6 months in 2019 and 2020. Before each campaign, a churn prediction model (independent of the models evaluated in this section) was trained on the whole customer base to predict the churn risk. The riskiest customers were randomly split into target and control groups. Customers in the target group were contacted by phone and were proposed a tariff plan adapted to the apparent root cause of potential churn. For example, if a large amount of mobile data was used, a tariff plan with a larger provision of mobile data was then suggested. The final data set used in this section comprises only customers selected in the target and control groups, all other customers that are not part of the campaign are discarded. The data set contains 11268 samples, for 145 features. Examples of features include the tariff plan of the customer, the number of calls over the last month, some socio-demographic information, the number of calls to customer service, and so on. The churn rate in the control group is 4.85\%, while in the target group it is 4.03\%. The control group amounts to 33\% of the data set. Note that the treatment indicator in this data set indicates whether a call attempt to the customer was made, and does not indicate whether the customer answered the call or accepted the offer.

\subsection{Methodology}
We train an uplift random forest model~\citep{guelman2015uplift} on the O. data set using the R package \emph{uplift}~\citep{guelman2014package}. Other uplift models have been shown to be superior in accuracy (e.g. the X-learner~\citep{kunzel2019metalearners}) but we need here separate estimators for $S_0(x)$ and $S_1(x)$ to compute the uplift bounds and the point estimator. This condition is satisfied by the uplift random forest, as well as the T-learner approach~\citep{kunzel2019metalearners}. The uplift random forest model is trained with 100 trees. Given the high imbalance of the data sets, we rely on the EasyEnsemble strategy~\citep{liu2009exploratory} for class balancing. It consists in training $k$ base learners ($k=10$ in our case) on the whole
set of positive instances (churners) and an equally sized random set of negative instances. This choice is based on previous literature on similar tasks with high imbalance and large class overlap~\citep{zhu2017empirical,dalpozzolo2014learned}. The predictions of all the base learners are averaged to obtain the final prediction. When a resampling strategy such as EasyEnsemble is used to obtain a balanced data set, the prior probability of churn is modified~\citep{batista2004study}, and the scores predicted by the trained model are biased. This bias is corrected with the calibration formula presented by \cite{dalpozzolo2015calibrating}. To avoid overfitting on a specific train-test split, we repeat the experiment using a k-fold cross-validation scheme with $k=5$.

\begin{figure}
    \vskip 0.2in
    \begin{center}
    \centerline{\includegraphics[width=0.8\columnwidth]{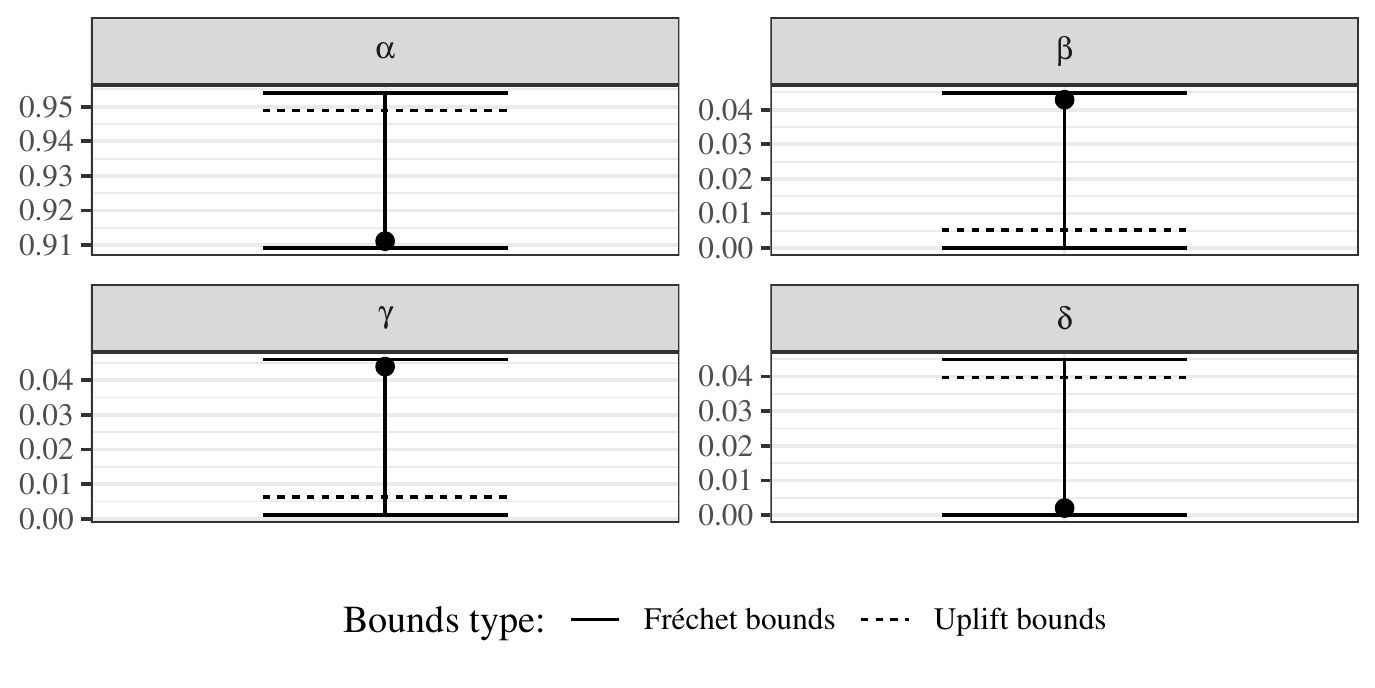}}
    \caption{Point estimate and bounds on $\alpha,\dots,\delta$. Note the different vertical axis for $\alpha$.}
    \label{fig:emp_beta_bounds}
    \end{center}
    \vskip -0.2in
\end{figure}

\subsection{Results}
The estimated distribution of counterfactuals is reported in Figure~\ref{fig:emp_beta_bounds} and Table~\ref{tab:bounds_emp}. In Figure~\ref{fig:emp_beta_bounds}, each of $\hat\alpha$, $\hat\beta$, $\hat\gamma$ and $\hat\delta$ is reported in a different sub-plot, together with the uplift and Fréchet bounds. We observe that the uplift bounds are consistently tighter than the Fréchet bounds, although not by a large margin. The value of $\hat\beta$ and $\hat\gamma$ are very close, with point estimates at respectively $4.29\%$ and $4.39\%$. The value of $\hat\alpha$ is high, around $91.12\%$, as expected since most customers do not churn.

\begin{table}
    \begin{center}
    \begin{minipage}{\textwidth}
    \caption{Numerical values of the estimated counterfactual distribution $\alpha,\dots,\delta$ on the O. data set. The uplift bounds and the Fréchet bounds show similar results.}\label{tab:bounds_emp}%
    \centering
    \begin{tabular}{lcccc}
        \toprule
        & $\alpha$ & $\beta$ & $\gamma$ & $\delta$ \\
        \midrule
        Point estimate &  $91.12$ & $4.29$ & $4.39$ & $0.20$ \\
        Uplift bounds      &  $[90.91,94.89]$ & $[0.52,4.49]$ & $[0.62,4.60]$ & $[0.00,3.98]$ \\
        Fréchet bounds &  $[90.91,95.40]$ & $[0.00,4.49]$ & $[0.11,4.60]$ & $[0.00,4.49]$ \\
        \bottomrule
    \end{tabular}
    \end{minipage}
    \end{center}
\end{table}

\begin{figure}
    \vskip 0.2in
    \begin{center}
    \centerline{\includegraphics[width=0.7\columnwidth]{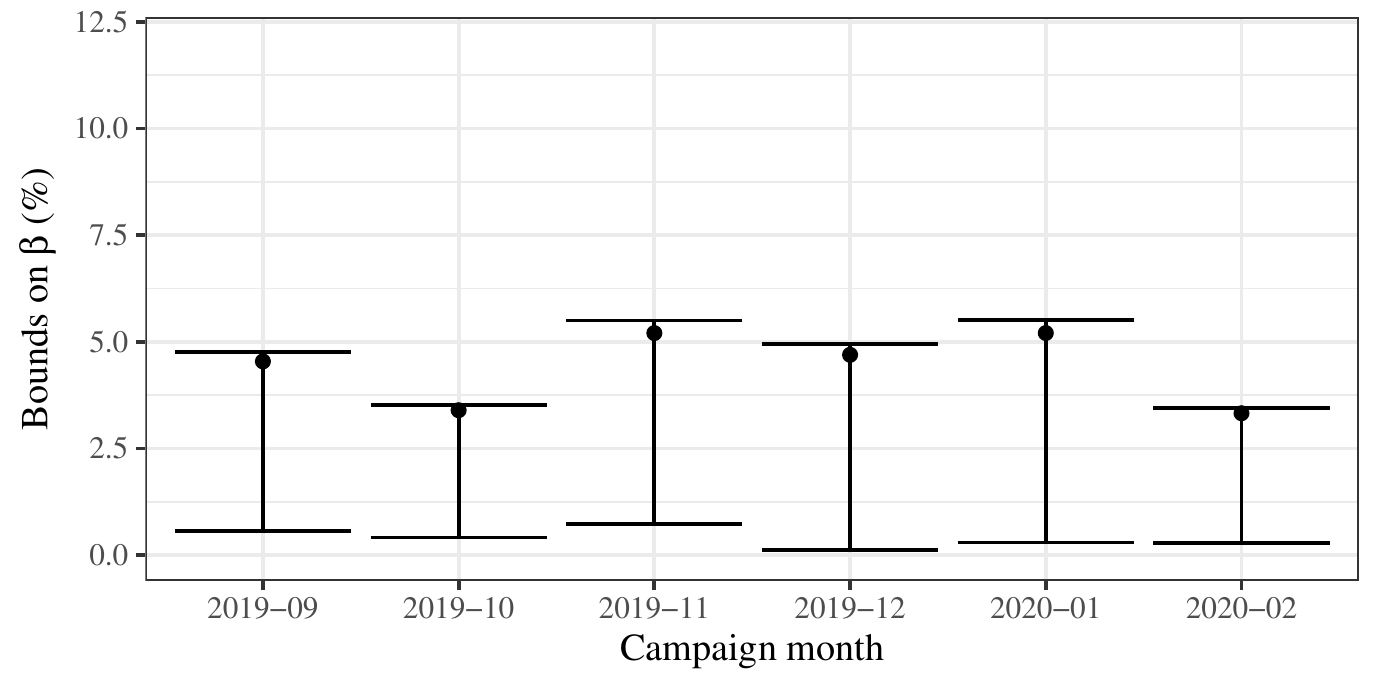}}
    \caption{Point estimate and uplift bounds on $\beta$, for each month of the campaign.}
    \label{fig:emp_beta_bounds_months}
    \end{center}
    \vskip -0.2in
\end{figure}

The proportion of persuadable customers is estimated as $\hat\beta=4.29\%$, with a lower bound of $0.52\%$ and an upper bound of $4.49\%$. This amounts to 483 customers, bounded between 58 and 505. This indicates that a maximum of 505 customers should have been called during the 6-months campaign, while in practice 7500 customers have been called. We applied the same methodology separately for each month instead of on the whole campaign data, and the results are reported in Figure~\ref{fig:emp_beta_bounds_months}. We observe that, although the value of $\hat\beta$ seems to fluctuate from one month to the next, it tends to be close to the upper bound. This is because both $\widehat S_0(x)$ and $\widehat S_1(x)$ tend to be close to zero, and $\hat\beta(x)$ is estimated as $\widehat S_0(x)(1-\widehat S_1(x))$ in Equation~\eqref{eq:beta_hat}. Therefore $\hat\beta(x)$ is typically close to $\widehat S_0(x)$, and the upper bound $\min\{\widehat S_0(x), 1-\widehat S_1(x)\}$ from Equation~\eqref{eq:bounds_b} is almost always equal to $\widehat S_0(x)$ as well.

\subsection{Profit analysis}
To give some intuition about these results, we now conduct a simplistic profit analysis. Let us suppose that each call has a cost $C=1\text{\euro}$, and that the average customer lifetime value is $V=120\text{\euro}$ (a customer pays on average 20\euro\,per month and stays 6 months). The benefit due to the campaign as it actually happened can be computed as
\begin{equation}
    \label{eq:profit}
    \text{Profit}=NUV - NC
\end{equation}
Where $N$ is the number of contacted customers and $U=S_0-S_1$ is the campaign uplift (approximately $0.8\%$ in our case). The term $NUV$ is the benefit generated by converting customers. The benefit of calling do-not-disturb customers cancels out the benefit of calling persuadable customers, since $U=\beta-\gamma$.\footnote{This can be shown by decomposing $U=P(Y_0=1)-P(Y_1=1)$ in terms of $\beta,\gamma$ and $\delta$.} The term $NC$ in Equation~\eqref{eq:profit} is the cost of calling $N$ customers. By evaluating this expression on the O. data set, we obtain that the campaign incurred a net loss of 130\euro. However, if we suppose that we were able to call only the 483 persuadable customers, the campaign could generate a profit of up to 57477\euro. Note that this is a simplistic way to evaluate the profit generated by a campaign. For more detailed estimations of the profit, we refer the reader to~\citep{li2019unit,verbraken2013novel,verbeke2012new,gubela2021uplift}.

\subsection{Discussion}
The improvement of the uplift bounds with respect to the Fréchet bounds is directly related to the quantity of information between the features and the outcome (see Theorem~\ref{thm:bounds_span}). The small improvement observed in practice, as shown in Figure~\ref{fig:emp_beta_bounds}, indicates that the uplift terms, and in turn counterfactual probabilities, are difficult to estimate in real-world settings such as customer churn prediction. A possible solution would be to add more informative features or design a more powerful uplift model. The bounds can also be further refined when observational data is available (i.e. data where the treatment assignment is not randomized), as demonstrated in \citep{mueller2022personalized}. The results of this section provide nonetheless very valuable insights for our industrial partner O. on the potential value of past retention campaigns and on the distribution of the different customer categories.

The results of this section do not indicate which customers should be targeted in order to maximize the profit from the retention campaign. This is the objective of uplift modeling. There is some debate on whether uplift modeling is always the best approach for causal decision-making. \citet{fernandez-loria2022causal,fernandez-loria2022causala} show that uplift models are sub-optimal under some circumstances, and that proxy targets such as the probability of the outcome are sometimes more effective for accurate causal decision-making. This is in line with the abundant literature on churn management that use predictive models instead of uplift models, e.g. \citep{amin2019customer,coussement2017comparative,oskarsdottir2018time} to cite a few. \citet{li2019unit} consider the case where each of the four categories of customers (\emph{persuadable}, \emph{sure thing}, \emph{lost cause} and \emph{do-not-disturb}, see Table~\ref{tab:customer_categories}) have arbitrary associated costs. In this case, counterfactual identification is essential for accurate causal decision-making.

\section{Conclusion}
\label{sec:conclusion}
We have derived and empirically assessed new bounds and a point estimator on the probability of counterfactuals for binary outcomes under the assumption of unconfoundedness. Counterfactuals are essential for accurate decision-making for example in churn prevention in the telecom industry.

The proposed uplift bounds improve upon the classical Fréchet bounds by leveraging the scores estimated by an uplift model. We have demonstrated theoretically that the bounds improve as the quality of the uplift estimation increases. Simulated examples indicate that the uplift bounds typically provide a significant improvement over the Fréchet bounds. We have also derived a point estimator by assuming the conditional independence between the potential outcomes $Y_0$ and $Y_1$. Simulated examples demonstrate that the estimator is still close to the true value even when this condition is not respected.

Our estimators are limited by several factors. The most important is the choice of the underlying uplift model. The uplift model should be unbiased, and the quality of the estimator depends on the quality of the uplift model. Since the two uplift terms $S_0(x)$ and $S_1(x)$ are used independently in our estimators, we are also limited to uplift estimators that can provide an estimation of these two terms separately.

Counterfactuals model individual behavior and as such can provide significant business insights about customers. In future work, we intend to explore the relationship between counterfactuals and customer features. This will allow describing the persuadable customers in terms of concrete characteristics, which very desirable from a business standpoint.

\bibliography{references}

\end{document}